\documentclass[aps,prl,preprint,unsortedaddress,showkeys,nofootinbib]{revtex4-1}

\usepackage{amssymb,amsmath,amsfonts}

\usepackage{graphics}
\usepackage{graphicx}
\usepackage{epstopdf}
\usepackage[pdftex]{hyperref}
\usepackage{longtable}

\usepackage{bm}
\usepackage{subfigure}

\usepackage{color}
\usepackage{setspace}
\usepackage{multirow}

\usepackage[ruled,vlined,linesnumbered]{algorithm2e}

\usepackage{color, colortbl}
\definecolor{Yellow}{rgb}{1,1,0}
\definecolor{Grey}{rgb}{.87,.87,.87}
\definecolor{Purple}{rgb}{.8,.0,1.0}
\definecolor{Crimson}{rgb}{.86,.08,.23}

\begin{document}

\title{Hierarchical Graph Neural Networks}

\author{Stanislav Sobolevsky
\footnote{To whom correspondence should be
addressed: sobolevsky@nyu.edu}}
\affiliation{
Center For Urban Science+Progress, New York University, Brooklyn, NY, USA\\
}



\date{\today}

\begin{abstract}
\begin{it}
Over the recent years, Graph Neural Networks have become increasingly popular in network analytic and beyond. With that, their architecture noticeable diverges from the classical multi-layered hierarchical organization of the traditional neural networks. At the same time, many conventional approaches in network science efficiently utilize the hierarchical approaches to account for the hierarchical organization of the networks, and recent works emphasize their critical importance. This paper aims to connect the dots between the traditional Neural Network and the Graph Neural Network architectures as well as the network science approaches, harnessing the power of the hierarchical network organization. A Hierarchical Graph Neural Network architecture is proposed, supplementing the original input network layer with the hierarchy of auxiliary network layers and organizing the computational scheme updating the node features through both - horizontal network connections within each layer as well as the vertical connection between the layers. It enables simultaneous learning of the individual node features along with the aggregated network features at variable resolution and uses them to improve the convergence and stability of the individual node feature learning. The proposed Hierarchical Graph Neural network architecture is successfully evaluated on the network embedding and modeling as well as network classification, node labeling, and community tasks and demonstrates increased efficiency in those.
\end{it}
\end{abstract}


\maketitle

\section*{Introduction}

The recent decade has seen explosive growth in the number of deep learning applications in all areas of machine learning, as the computational power and learning techniques have matched the complexity of the deep neural network (NN) architectures needed for successful applications. And just like vanilla and convolutional neural networks (CNN) successfully handle supervised classification and unsupervised pattern detection in vectorized data or arrays (like 2d or 3d images), the graph neural networks (GNN) in their recent interpretation \cite{defferrard2016convolutional, kipf2016semi} are used for the networked data. Supervised classifications and unsupervised embedding of the graph nodes find diverse applications in text classification, recommendation system, traffic prediction, computer vision, etc. \cite{wu2020comprehensive}. Traditional complex network analysis problems like network community detection have also seen successful applications of the GNNs and similar techniques \cite{bruna2017community, shchur2019overlapping, bandyopadhyay2020self, sobolevsky2021recurrent}.

Traditional NNs supplement the input layer with a hierarchy of the hidden layers, which eventually lead to the output layer, having the weights of each neural connection trained independently.
In contrast with the traditional NNs, the GNNs effectively consist of a single layer of neurons, serving as both - the input and the output layer - and implement recurrent computation propagation through it. With that, all the neurons (network nodes) are trained to have the same computational parameters across the entire network. Those parameters, however, may in some configurations, change between computational iterations. This allows imagining having multiple computational layers, although the topology and learnable neuron weights within each layer stay the same (e.g. a now popular two-layer convolutional GNN \cite{kipf2016semi}).

The computational scheme of the GNNs is leveraging the network topology by accounting for the weights input from the neighbors (either by original network edge weights or by some modification of those, e.g., through the weights in the Laplacian matrix). However, apart from the edge weights, all the neighbor nodes are treated equally. At the same time, a recent paper \cite{grauwin2017identifying} emphasizes the critical importance of accounting for the hierarchical organization in modeling, for example, the networks of human mobility and interaction. And such networks often demonstrate prominent community and hierarchical organization at city \cite{kang2013exploring, sobolevsky2018twitter}, country \cite{Ratti2010GB, Sobolevsky2013delineating, amini2014impact, sobolevsky2014money} or global scale \cite{hawelka2014geo, belyi2017global}.

The ideas of leveraging different kinds of hierarchical architecture within recurrent \cite{dai2017deeptrend} or convolutional \cite{roy2020tree} NNs, accounting for possibly hierarchical hypergraph organization within the input layer of GNNs \cite{feng2019hypergraph}, and introducing various hierarchical approaches within GNNs \cite{wu2020learning, gao2021ipool, xu2020global, nassar2018hierarchical, mo2019structurenet} have been over the few recent years. Most of them use some particular approaches addressing specific applied problems using special types of pre-defined network hierarchies, associated with those problems (e.g. road segments and two levels of their groupings in \cite{wu2020learning}). And although the term "hierarchical graph neural networks" starts being actively used, the common consensus on its meaning is yet to emerge.

At the same time, the idea of supplementing an initial graph input layer of the GNN with a hierarchy of auxiliary hidden layers like in traditional NNs allows an intuitive generalization useful for a variety of possible applications. This paper aims to define such a general model and build a mathematical framework for it, aiming for broad applicability.

The proposed Hierarchical Graph Neural Network (HGNN) architecture, in a sense, connects the dots between the traditional NNs, GNNs, and the network analysis. It leverages the hierarchical structure of the network to supplement the original graph input layer with the hierarchy of additional hidden layers of neurons connected accordingly within themselves and with the original network nodes. In particular, the layers could be seen as performing some sorts of network aggregations along the network hierarchical structure. And the computational flow within HGNN recurrently updates the node features running in both dimensions – horizontal within each layer (similarly to the GNNs) and vertical across the layers (just like in a traditional NN). This way, the node features will get updated based on the neighbor nodes within the same network layer as well as the features of the connected nodes from the neighbor layers. 

This architecture enhances the learning of the individual node features with aggregated network features, simultaneously learned at different resolutions. It will improve the overall stability of the model, optimize the cumulative dimensionality of the node features to be learned across the layers. It will also introduce additional flexibility into the computational scheme by adding further types of neurons at each hierarchical layer with different sets of trainable weights.

\begin{figure}
    \centering
    \includegraphics[width=0.7\textwidth]{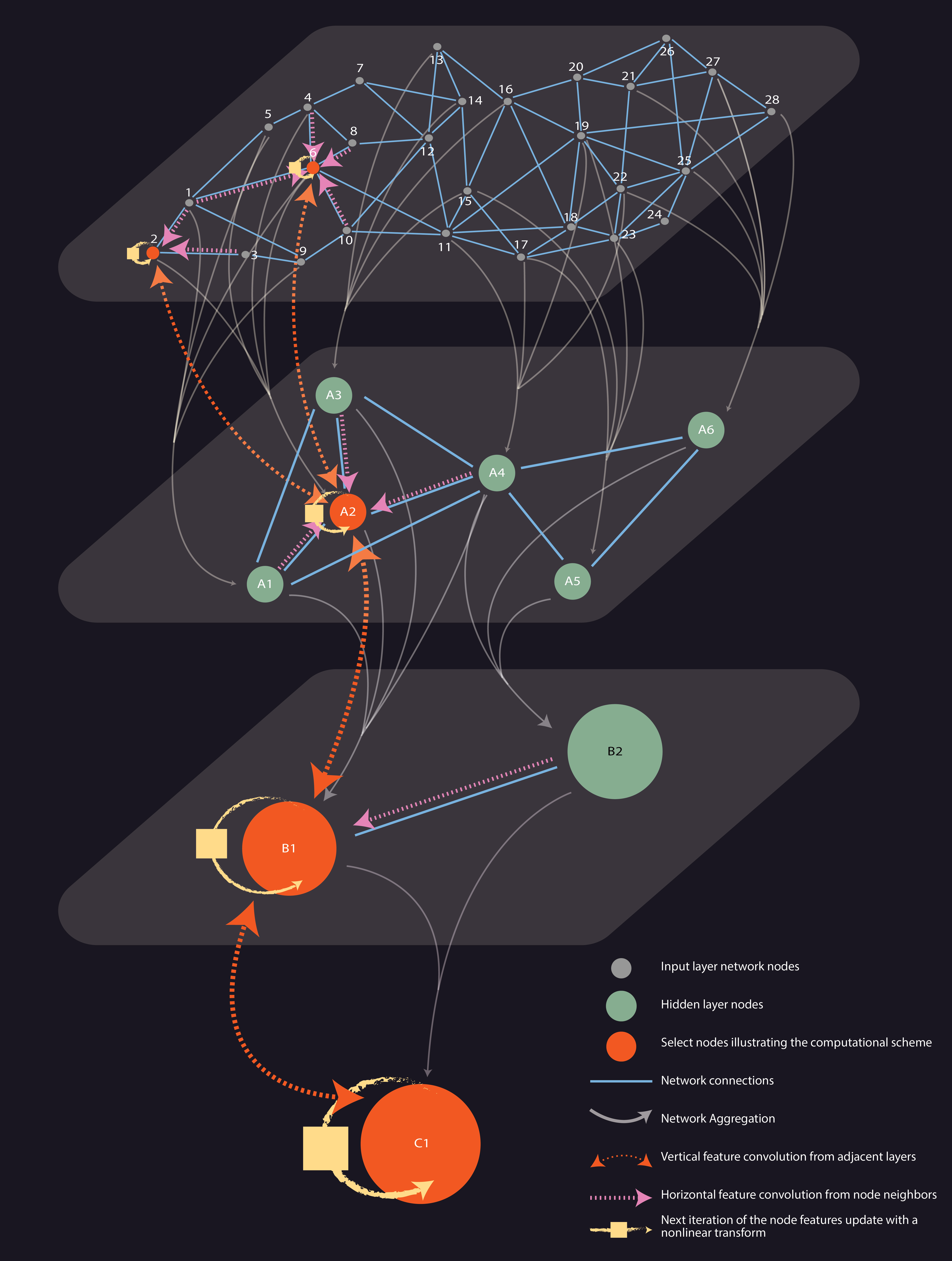}
    \caption{HGNN architecture supplementing the input network layer with a hierarchy of hidden layers and running a computational flow recurrently updating each node features in both dimensions - within the layers and across the layers}
    \label{fig:HGNN}
\end{figure}

\section{The Hierarchical Graph Neural Network Approach}

The proposed Hierarchical Graph Neural Network (HGNN) architecture (fig. \ref{fig:HGNN}) supplements the original input network layer $(L^0, A^0)$ with a hierarchy of hidden network layers $(L^h, A^h), h=\overline{1..h^*}$, and connections between the neighbor layers $H^{h-1\to h}$ and $H^{h+1\to h}$. Here the $L^h$ denote the sets of nodes at each layer $h$, while $A^h$ denote the layers' weighted adjacency matrices. The $H^{h-1\to h}$ or $H^{h+1\to h}$ denote the adjacency matrices of the bipartite networks, incorporating the connection weights between the nodes of the layers $h-1$ or $h+1$ and the nodes of the layer $h$. The goal is to learn the feature vectors $x^h(a^h)$ of the given dimensionality characterizing each network node $a^h\in L^h$ according to the selected model objectives. The main learning target corresponding to node classification or network embedding may include features for the individual nodes $a^0\in L^0$ of the original input layer. While sometimes, e.g., in the case of network classification, the objective might be to learn the holistic network features associated with one or several nodes of the last layer $L^{h^*}$, thus serving as an output layer. 

The computational flow learning the HGNN features starts with initial feature assignment $x^h_0(a^h)$ for each node $a^h$ at each layer $h$, and iterates to recurrently update each node features in both dimensions: horizontal flow at each layer updating the node features based on the neighbor nodes’ features at this same layer, and vertical, accounting for the features of the connected nodes from the neighbor layers. Each node updates its state based on the information aggregated from its connections as well as its current state:
\begin{equation}
\begin{array}{c}
x^h_{i+1}(a^h)=f^h_i\left(x^h_i(a^h),\ \sum_{b^h\in L^h} \tilde{A}^h(b^h, a^h) x^h_i(b^h),
\right.\\
\left.
\sum_{b^{h-1}\in L^{h-1}} H^{h-1\to h}(b^{h-1}, a^h)x^{h-1}_i(b^{h-1}), 
\right.\\
\left.
\sum_{b^{h+1}\in L^{h+1}} H^{h+1\to h}(b^{h+1}, a^h)x^{h+1}_i(b^{h+1})\right),
\ i=\overline{0..I-1}
\end{array}
\label{eq:HGNN}
\end{equation}
where $\tilde{A}^h$ are certain derivative matrices characterizing the network connections at each hierarchical level $h$ (like normalized adjacency $(D^h)^{-1/2}A^h (D^h)^{-1/2}$ or normalized Laplacian $(D^h)^{-1/2} (D^h-A^h) (D^h)^{-1/2}$ where $D^h$ is the diagonal matrix of the node weighted out-degrees or in-degrees). The nonlinear activation functions $f^h_i$ with trainable parameters could be implemented as the generalized linear functions applying a nonlinear one-dimensional or low-dimensional transform to a linear combination of the input variables or, more generally, as a shallow fully connected vanilla neural network with trainable weight parameters.

The parameters of $f^h_i$ can stay constant over all the iterations $i$ ($f^h_i=f^h$ implementing a recurrent HGNN) or can vary over a limited number of iterations (shallow HGNN).

The hierarchical structure of the HGNN is defined through the edge weights $H^{h-1\to h}$ and $H^{h+1\to h}$ as well as the network connections $A^h$ for $h>0$, that could be either predefined or included among other features to be learned by the HGNN. 

The final output of the model (\ref{eq:HGNN}) could be the features $x^0_I(a^0)$ for each of the nodes $a^0\in L$ of the original network learned on the last iteration $I$ or a certain final functional transform of those $g(x^0_I(a^0))$. Alternatively, e.g., if the model is used for the holistic network classification, then the output could be the features $x^{h^*}_I(a^{h^*})$ or the transformed features  $g(x^{h^*}_I(a^{h^*}))$ for one or several nodes $a^{h^*}\in L^{h^*}$ of the top $h^*$-th layer of the hierarchy, which thus serves as an output layer. The function $g$ implementing a final transform of the output features could range from simply taking a subset of the output features or being a generalized linear function with trainable weight parameters, all the way to being a vanilla neural network of an arbitrary level of complexity.

\section{Defining the hidden layers}

The neurons of the hidden layers can perform nested network aggregation over the groups of nodes from the previous layers following the inferred hierarchical structure of the network. E.g. define a nested hierarchy of network discrete partitions $c^h:L^h\to \{0,1\}^{|L^{h+1}|}$ or probabilistic partitions $c^h:L^h\to [0,1]^{|L_{h+1}|}$ at each level $h$, including the initial level $h=0$, such that $\forall a^h\in L^h, \| c^h(a)\|_1=1$.
This way components of the vector $c^h(a^h)$ represent probabilities of attaching the node $a^h\in L^h$ to each node $a^{h+1}\in L^{h+1}$.
Partitions can also be represented as $|L^h|\times |L^{h+1}|$-matrices $C^h=(c^h(a^h): a^h\in L^h)$ stacking all the vectors $c^h(a^h)$ together.
Then for each $h$ define the $h+1$-th hierarchical network layer as an aggregation $A^{h+1}:=(C^h)^T A^h C^h$ of the previous layer $A^h$ with respect to the selected discrete or probabilistic partition $C^h$. 

For defining the inter-layer connections $H^{h\to h\pm 1}$ effectively implementing feature aggregation/disaggregation, we shall also need to define the relative weights $v^0(a^0)>0$ defined for all the nodes $a^0$ of the initial input layer $L^0$. They can be further aggregated to weight the nodes of the further layers using a recurrent relation $v^{h+1}:=(C^h)^T v^h$. 

Then the connections $H^{h\to h+1}$ and $H^{h+1\to h}$ between the layers $h$ and $h+1$ could be defined in one of the two ways:\\ 
a) weighted averaging $H^{h\to h+1}(a^h,a^{h+1}):=\frac{C^h(a^h,a^{h+1})v^h(a^h)}{\sum_{b^h\in L^h}C^h(b^h,a^{h+1})v^h(b^h)}$ to aggregate the node features for their next layer attachments, and reverse averaging $H^{h+1\to h}:=(C^h)^T$ aiming to reconstruct the node features based on their next level attachments; such averaging would be appropriate if the node features represent some sort of normalized quantities like, e.g., probabilities of a binary node labeling or node community attachment;\\
b) feature additive aggregation $H^{h\to h+1}:=C^h$ and weighted disaggregation\linebreak $H^{h+1\to h}(a^{h+1},a^h):=\frac{C^h(a^h,a^{h+1})v^h(a^h)}{\sum_{b^h\in L^h}C^h(b^h,a^{h+1})v^h(b^h)}$, suitable if the node features represent cumulative quantities of some sort for which an additive aggregation would be appropriate. 

Either way ensures $H^{h\pm 1\to h}(a^{h\pm 1},a^h)\in [0,1]$ with appropriate normalization on one of the sides: for a)  $\sum_{a^h\in L^h}H^{h\pm 1\to h}(a^{h\pm1},a^h)=1$, and for b) $\sum_{a^h\in L^h}H^{h\to h\pm 1}(a^h,a^{h\pm 1})=1$.

With the network connections $A^h$, defined within each aggregated network layer and the aggregation/disaggregation connections $H^{h\to h+1}$ and $H^{h+1\to h}$ defined between the layers, the HGNN can run the computations recurrently updating the node features at each aggregation layer based on the neighbor node features as well as the associated nodes at the lower and higher aggregation levels. This way, the synchronous feature learning processes over different layers will aid each other, e.g., features learned for the nodes' hierarchical aggregations at various resolutions will supplement learning of the individual node features, improving the overall stability and reducing the dimensionality of the model.

The nested hierarchical mappings $c^h$ could be pre-defined through discrete network community/sub-community detection \cite{combo, Sobolevsky2013delineating}, probabilistic partitions \cite{sobolevsky2021recurrent}, or hierarchical inference \cite{sobolevsky2017inferring}. While the relative weights $v^0$ could be defined, for example, based on the nodes' weighted degrees.

On the other hand, the probabilistic attachments $c^h$ of the nodes $a^h\in L^h$ of the given dimensionality $|L^{h+1}|$ could be learned as part of the node feature vectors $x^h_i$ at each hierarchical level $h$. And the relative weights $v^0$ could be learned as parts of the $x^0_i$. They could be initialized randomly or through initial network hierarchical inference as above and then recurrently updated at each iteration $i$. While the hierarchical structure used at the iteration $i+1$ will be based on the sub-vectors $c^h_i\subset x^h_i$ learned on the previous iteration. The constraints $c^h_i\in [0,1]$ could ensured by using appropriate limited activation functions $f^h$, like sigmoid or rectified linear unit functions, while appropriate regularization or normalization could ensure $||c^h_i||_1=1$. 

Graph neural network-type architectures were already proven to be applicable for learning node community attachments \cite{sobolevsky2021recurrent}. And  HGNN looks like a proper generalization to implement a hierarchy of nested node attachments. This way, it could simultaneously achieve two goals within the same learning process: infer a suitable auxiliary network hierarchical representation, and apply it to learn the node features according to the given model's objective. Or, if learning the network hierarchical representation is the primary goal, the model could aim at optimizing its quality, quantified with an appropriate quality function.

\section{Training the HGNN}

The trainable parameters of the HGNN include weight parameters of all the activation functions $f^h_i$ or $f^h$ in case of a recurrent HGNN. The appropriate quality function to assess the final HGNN's performance over a given input (which consists of one or several networks to initialize the input layer) is chosen as an objective function. Then for relatively shallow architectures with a small $I$, the parameter training could be implemented using standard backpropagation approaches. While for recurrent HGNN with higher $I$ a genetic optimization or Neuroevolution \cite{stanley2019designing} training could be preferable.  

Depending on the purpose of the HGNN, different objective functions could be used to train the model:

\begin{itemize}
  \item a loss function like mean squared error or binary cross-entropy between the learned node output features $g(x^0_I)$ and the ground-truth labels $x^*$ in case of a supervised node labeling;
  \item a reconstruction loss in case of learning the network embedding aiming for network reconstruction (see below);
  \item a custom network quality function based on $g(x^0_I)$ in case of unsupervised node labeling, e.g. network modularity in case of learning the node community attachment, like in \cite{sobolevsky2021recurrent};
  \item network classification accuracy quantified by comparing the output classification labels $g(x^{h^*}_I)$ and the ground-truth labels $x^*$;
  \item a custom quality function based on $g(x^{h^*}_I)$ in case of unsupervised network labeling, e.g. discovering common categories of network topology.
\end{itemize}

In case of the network embedding problem for example, the objective could be to learn the paired $d$-dimensional feature vectors $x^0(a^0)=(l(a^0), r(a^0))$ for each node $a^0\in L^0$, that would allow to model the directed edge weights from $A^0$ to a certain degree of accuracy. Consider a general probabilistic model representing an edge weight $w(a,b)$ between a pair of nodes $a$ and $b$ as
$$
w(a,b)\sim U(l(a),r(b))
$$
where $U$ is a random variable with the distribution defined by the features $l,r$. The density function $p(w|l,r)$ of the random variable $U$ could be either pre-defined as a function of $l,r$ or represented with a certain functional form with trainable parameters, e.g. with a vanilla neural network with the inputs l,r and w. 

Then the objective of training the embedding model is to learn the $l(a^0), r(a^0)\in R^d$ for each node $a^0\in L^0$ as well as the learnable parameters of the density function $p(w|l,r)$, such that
the joint log-likelihood of the observed edge weights is maximized:
\begin{equation}
{\cal L}=\sum_{a,b\in L^0} \ln\left(p(w(a,b)|l(a),r(b)\right)\to \max
\label{logL}
\end{equation}

For a non-weighted network a Bernoulli model could be appropriate, where
$$
P(w(a,b)=1)=p(l(a),r(b)),
$$
where the probability $p$ of having an edge is a certain function of $l,r$. This function could be pre-defined (e.g. $p(l,r)=\sigma(l^T\cdot r)$ where $\sigma$ denotes a sigmoid function) or learned during the model training. Then the objective log-likelihood function (\ref{logL}) to maximize will then be a standard binary cross-entropy
\begin{equation}
{\cal L}=\sum_{a,b\in L^0} \left[w(a,b) \ln p(l(a),r(b)) + (1-w(a,b))\ln (1-p(l(a),r(b))\right].
\label{logL_binary}
\end{equation}

For a weighted network, where edge weights represent certain aggregated  quantities, like volumes of human activity between locations, a Gaussian model could be considered, where
$$
w(a,b)\sim N(\mu(l(a),r(b)),\sigma(l(a),r(b))),
$$
while the mean $\mu$ and the standard deviation $\sigma$ of the distribution are certain functions of $l,r$. Again, those functions could be pre-defined or learnable. Then the objective function (\ref{logL}) will take a streightforward analytic form
$$
{\cal L}=-\sum_{a,b\in V} \frac{(w(a,b)-\mu(l(a),r(b))^2}{2\sigma(l(a),r(b))^2}-\sum_{a,b\in V}\ln(\sigma(l(a),r(b)))+\ldots
$$
up to an additive constant, not affecting the optimization.

A case with a constant $\sigma$ leads to an objective of minimizing the cumulative squared error between the actual weights and the model estimates $\mu(l(a),l(b))$, namely
$$
\sum_{a,b\in V} (w(a,b)-\mu(l(a),r(b)))^2\to\min
$$

In particular, a dot-product form for $\mu=l(a)^T \cdot r(b)$ with a constant $\sigma$ (i.e. a squared error objective function) would correspond to a well-known standard embedding based on a singular vector decomposition (SVD) with $r$ defined as the $d$ leading right singular vectors, and $l$ as $d$ leading left singular vectors, both scaled by the square roots of singular values.

However in many cases it would be more appropriate to fit the $\sigma$ along with $\mu$ as the estimates for the higher weight edges would naturally have higher errors. Also for networks with non-negative edge weights one might want to ensure that $\mu$ is positive. So a standard SVD embedding has significant theoretic limitations. 

And a HGNN model will aim at constructing the final embedding $\tilde{x}^0(a^0)=(l(a^0), r(a^0))$ for all the nodes $a^0\in L^0$, by combining the linear or nonlinear combination of the features $x^0(a^0)$ directly learned for the node $a^0$ as well as the features learned for the connected nodes at various hidden layers
\begin{equation}
\tilde{x}^0(a^0)=\Theta\left(x^0(a^0), \sum_{b^h\in L^h} H^{h\to 0}(b^h, a^0)x^h(b^h), h=\overline{1..H}\right),
\label{eq:titleX}
\end{equation}
where $H^{h\to 0} = \prod_{k=1}^{h}H^{k\to k-1}$ represent the indirect connections between the layer $L^h$ and the input layer $L^0$ and $g$ is a linear or non-linear transform. The embedding will provide a $2d\times n$ dimensional representation, enabling the efficient reconstruction of the original $n\times n$ dimensional network adjacency matrix $A^0$, while maximising the reconstruction quality (\ref{logL_binary}). The hierarchy of the hidden layers $(L^1, A^1), (L^2, A^2), \ldots$ (e.g., implementing nested network aggregations along with a suitable pre-defined or learned hierarchical representation) is introduced to reduce the overall dimensionality of the model. Indeed, the effective dimensionality of the final embedding $\tilde{x}^0(a^0)$ could include the dimensionality of the $x^0$, as well as all the $x^h$ learned for the hidden layers. While the total number of nodes within the hidden layers could be lower compared to the number of nodes of the original input layer.

In turn, learning the hierarchical structure of the highest utility for the network embedding HGNN could provide a meaningful network hierarchical representation, solving one of the network analysis problems.

\section{Initial proofs}

Initial utility proof of using aggregated features of the network communities while modeling dynamics of the temporal network and detecting anomalies in it is provided in \cite{he2019pattern}.

In order to evaluate the utility of the hierarchical representations within the network embedding framework above, we experimented with modeling the aggregated daily networks of taxi ridership over the year 2017, based on the open data provided by the NYC Taxi and Limousine Commission (TLC). We learned the hierarchical network representation for the three hierarchical levels $h=1,2,3$ using nested community detection as suggested in \cite{Sobolevsky2013delineating} implemented by a COMBO algorithm \cite{combo} over the historical 2016 data. And we use the above approach b) to define the inter-layer connections $H^{h\to h\pm 1}$ using the cumulative incoming mobility flows during 2016 as the relative node weights in the definition of $H^{h\to h-1}$.

\begin{figure}
    \centering
    \includegraphics{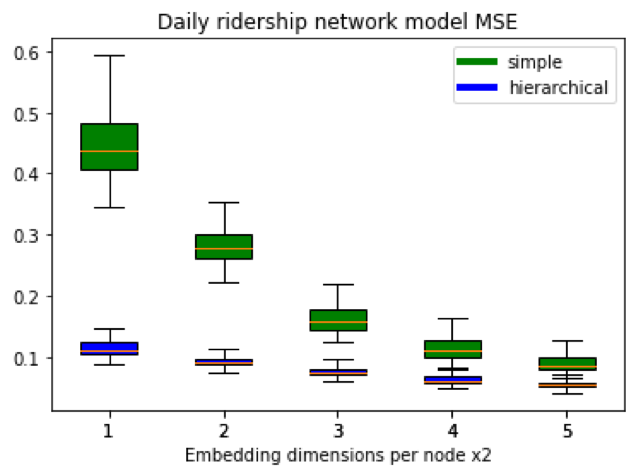}
    \caption{Performance of the simple single-layer and the hierarchical embedding models in terms of the daily MSE distributions (box plots) versus the effective dimensionality of the embedding to be learned per node}
    \label{fig:EmbeddingProof}
\end{figure}

At each hierarchical level we use the Gaussian embedding model with fixed $\sigma$ fitted using singular value decomposition or a GNN model leading to approximately the same results. While for (\ref{eq:titleX}) we use simple union $\tilde{x}^0(a^0)=\left(x^0(a^0), H^{h\to 0}(c^h(a^0), a^0)x^h(c^h(a^0)),\ h=1,2,3\right)$ where $c^h(a^0)$ denotes the community attachment of the node $a^0$ at the hierarchical level $h$. Then the efficient dimensionality of the learned embedding per node is computed as the cumulative dimensionality of all the features $x^h(a^h)$ learned over the entire HGNN, normalized by the size of the input layer $|L_0|$. As shown in Fig. \ref{fig:EmbeddingProof}, the hierarchical model achieves the same level of the normalized mean standard error (MSE, corresponding to the log-likelihood of the embedding model) of around 10\% for the daily ridership modeling as the non-hierarchical model with around four times smaller efficient dimensionality of the embedding to be learned (2 versus 8 dimensions per node one would need needed for the single-layer network embedding model).

\section{Conclusions}
The paper provided a mathematical framework for the novel Hierarchical Graph Neural Network Model. It connects the dots between the traditional NNs, GNNs, and the network analysis by leveraging the hierarchical structure of the network to supplement the original graph input layer with the hierarchy of additional hidden layers of neurons. The layers are connected with each other like in traditional NNs as well as internally within themselves like in GNNs. 

The proposed architecture enhances the learning of the individual node features with aggregated network features, simultaneously learned at different resolutions. It is meant to improve the overall stability of the model. It may also decrease the cumulative dimensionality of the node features to be learned to achieve the desired model performance. It also introduces additional flexibility into the computational scheme by adding different types of neurons at each hierarchical layer with different sets of trainable weights.

The model is suitable for a broad variety of machine learning problems over the networked input data: supervised and unsupervised node labeling, including network community detection, node embedding for network modeling, network supervised and unsupervised classification. The practical evaluation of the model performance in a variety of such applications is subject to future work.

\section{Acknowledgements}
The author thanks Mingyi He from New York University for her valuable help with the graphics. The author further thanks Satish Ukkusuri and Vaneet Aggarwal from Purdue University and Sergey Malinchik from Lockheed Martin for stimulating discussions. 

\bibliographystyle{apsrev}
\bibliography{HierGNN}

\end{document}